\documentclass[]{fairmeta}

\title{\methodname{}: Credit Assignment for Long-Horizon Agentic RL via Intra-group Rescaling}

\author[1,2*]{Fangcong Yin}
\author[1,3]{Ming Yin}
\author[1]{Tanvi Bhandarkar}
\author[1]{Aashu Singh}
\author[1]{Eryk Helenowski}
\author[1]{Shlok Kumar Mishra}
\author[1]{Saghar Hosseini}
\author[1]{Karthik Abinav Sankararaman}
\author[2]{Greg Durrett}
\author[1]{Dinkar Juyal}

\affiliation[1]{Meta}
\affiliation[2]{New York University}
\affiliation[3]{Duke University}

\contribution[*]{Work done at Meta}

\usepackage{amsmath,amssymb}
\usepackage{mathtools}
\usepackage{array}
\usepackage{wrapfig}
\usepackage{xcolor}

\definecolor{light-purple}{RGB}{151,156,171}
\definecolor{blue-color}{RGB}{40,166,189}
\definecolor{pink-color}{RGB}{237,46,104}
\definecolor{dark-grey-color}{RGB}{79,91,102}
\definecolor{burntorange}{HTML}{BF5700}
\definecolor{selfcitepurple}{HTML}{7E57C2}   

\newtcolorbox{imagebox}[2][]{%
    enhanced,
    breakable,
    nobeforeafter,
    title={#2},
    coltitle=white,
    colframe=selfcitepurple!40!white,
    colback=selfcitepurple!3!white,
    colbacktitle=selfcitepurple!90!black,
    fonttitle=\bfseries,
    sharp corners,
    boxrule=0.5pt,
    boxsep=1pt,
    left=6pt,
    right=6pt,
    top=0pt,
    bottom=0pt,
    #1
}

\newtcolorbox{modelsection}[2][]{%
    enhanced,
    breakable,
    title={#2},
    nobeforeafter,
    coltitle=black,
    colframe=selfcitepurple!40!white,     
    colback=selfcitepurple!4!white,       
    colbacktitle=selfcitepurple!22!white, 
    fonttitle=\bfseries,
    sharp corners,
    boxrule=0.5pt,
    boxsep=2pt,
    left=6pt,
    right=6pt,
    #1
}

\newtcolorbox[list inside=prompt,auto counter,number within=section]{prompt}[1][]{
    colbacktitle=black!80,
    colframe=black!80,
    coltitle=white,
    fonttitle=\sffamily\bfseries,
    fontupper=\ttfamily,
    boxsep=5pt,
    left=0pt,
    right=0pt,
    top=0pt,
    bottom=0pt,
    boxrule=1pt,
    enhanced,
    breakable,
    skin first=enhanced,
    skin middle=enhanced,
    skin last=enhanced,
    #1,
}

\DeclareMathOperator*{\mean}{mean}
\DeclareMathOperator*{\std}{std}
\DeclareMathOperator{\clip}{clip}

\newcommand{\Adv}{\widehat{A}}
\newcommand{\methodname}{\textsc{Self-Credit}}

\date{\today}
\correspondence{Fangcong Yin at \email{fcyin@meta.com}}

\abstract{
LLM agents are commonly trained with RLVR, but the sparse nature of typical RLVR rewards makes credit assignment challenging, especially for long-horizon tasks. We propose \methodname{}, an end-to-end agentic RL method that creates dense supervision signals from sparse outcome rewards using \textbf{self-generated credit assignment via intra-group rescaling}. \methodname{} builds on group-based RL with three components. First, during its rollouts, the policy model generates summaries with citations to earlier turns containing critical evidence, which serve as self-generated credit. Second, we refine this model-generated credit with \emph{intra-group rescaling}, which compares cited actions across successful and failed rollouts for the same prompt. This approach uses group-level agreement and disagreement to remove undeserved credit and further emphasize credit assigned to critical actions. Third, \emph{advantage reweighting} uses the rescaled credit weights to update the advantages of the cited turns. On BrowseComp-Plus, \methodname{} improves over the base models by $37.0$ and $36.2$ points on Qwen3 8B and 32B respectively, and outperforms the strongest agentic RL baselines by $8.5$ and $5.6$ points, respectively. \methodname{} also improves over the strongest baselines by $4.9$ points on GAIA and $4.7$ points on xBench-DeepSearch. Analyses show that self-generated credit provides useful signals for selecting critical actions, and that intra-group rescaling, especially on failed rollouts, is important across model scales.
}

\begin{document}

\maketitle

\section{Introduction}
  \label{section:intro}

\begin{wrapfigure}{r}{0.6\textwidth}
    \centering
    \includegraphics[trim={0cm 5cm 7cm 5cm}, clip,width=0.8\linewidth]{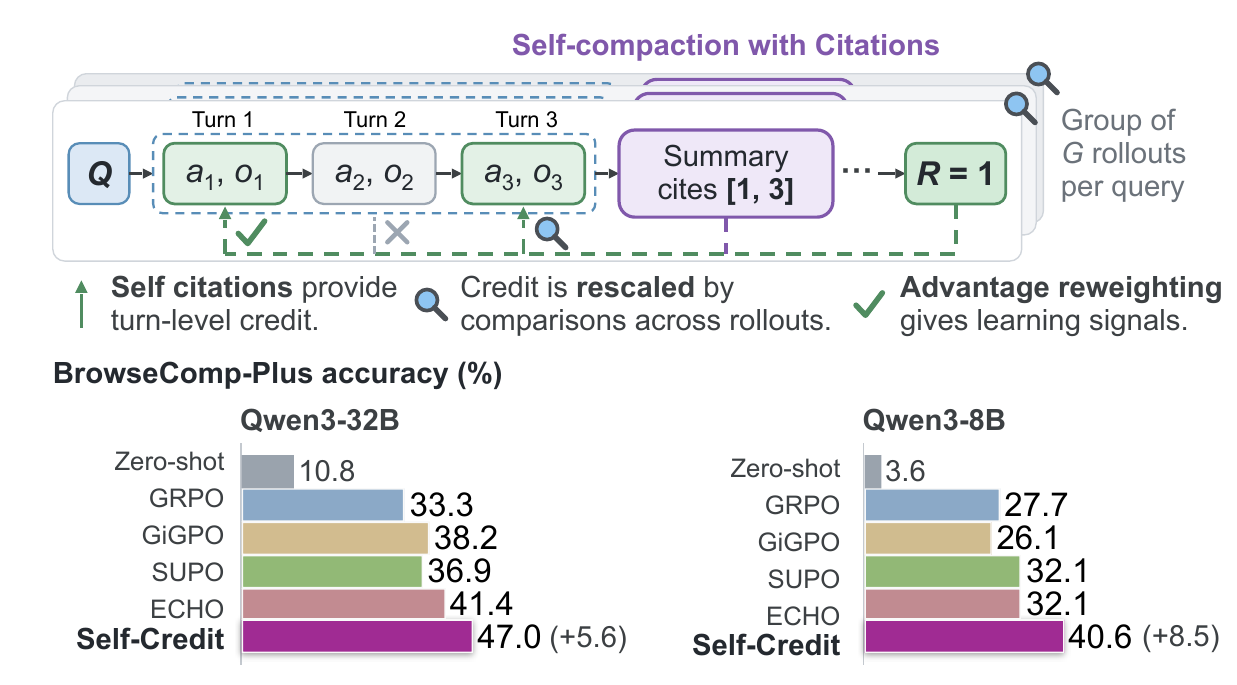}
    \caption{\methodname{} leverages self-generated summaries with Intra-group rescaling to provide credit assignment for long-horizon agentic RL. \methodname{} outperforms strong baselines on challenging long-horizon tasks, such as BrowseComp-Plus.}
    \label{fig:teaser}
\end{wrapfigure}

 Large language models (LLMs) are commonly used for agentic tasks, where they interact with environments and issue tool calls as actions. A principal challenge in building long-horizon agents is the nature of supervision: typically, models are trained via RLVR using only sparse rewards \citep{deepseek_r1}, as additional dense supervision is expensive to define and collect \citep{tao2025hybridreinforcementrewardsparse,setlur2025rewarding,zhang2025whichagent,zhang-etal-2025-lessons}. Recent work extends standard RLVR algorithms to provide credit assignment even when only sparse rewards are available, including approaches that use external judges \citep{zhang2026trca,zeng2025reinforcing} and learned critics \citep{zhang2025criticsearchfinegrainedcreditassignment,shi2026odysseusscalingvlms100}. However, these methods usually require additional supervision signals and often suffer from context growth in challenging long-horizon tasks \citep{sun2025scalinglonghorizonllmagent}.

Inspired by context management methods for LLM agents \citep{li2026compactionrl,xie2026echo}, we propose \methodname{}, a novel agentic RL method in which the policy LLM agent performs credit assignment through compaction using \textbf{self-generated credit with intra-group rescaling}. As illustrated in Figure~\ref{fig:teaser}, \methodname{} builds on a standard group-based RL algorithm with three components to create dense supervision signals from a single sparse outcome reward. First, we prompt the policy model to generate a summary with citations that identify earlier turns in the interaction history containing critical evidence for the agentic task. The summaries are used for compaction to compress the long context, and the citations serve as self-generated credit. Second, \emph{intra-group rescaling} refines this credit by comparing cited actions across successful and failed rollouts for the same prompt. In particular, this rescaling uses agreement and disagreement across the entire rollout group to determine how to remove undeserved credit and further emphasize credit assigned to critical actions, grounding the credit weights in group-level statistics. Third, \emph{advantage reweighting} uses the rescaled credit weights to selectively strengthen the advantages of tokens in model-cited turns to emphasize important behaviors in a long trajectory.

We evaluate \methodname{} on challenging long-horizon search and retrieval tasks, including multi-hop QA on BrowseComp-Plus and deep research on GAIA and xBench-DeepSearch. As shown in Figure~\ref{fig:teaser}, on BrowseComp-Plus, \methodname{} improves over the Qwen3-8B and Qwen3-32B base agents by $37.0$ and $36.2$ percentage points, respectively, and outperforms the strongest agentic RL baselines by $8.5$ and $5.6$ points, respectively. On deep research tasks, \methodname{} also outperforms the strongest evaluated baselines by $4.9$ points on GAIA and $4.7$ points on xBench-DeepSearch.

Ablations show that intra-group credit rescaling, especially downweighting repeated turns shared across failed rollouts, is important across model scales, and that using self-generated credit provides more effective credit assignment than the alternatives. Further analyses show that self-generated citations identify important actions related to gold answers, and that intra-group rescaling rules are applied selectively during training at a higher selection quality than baselines.

Our main contributions are as follows:
(1) We introduce \methodname{}, an end-to-end RL algorithm for training long-horizon agents using self-generated summaries for turn-level credit assignment with intra-group rescaling.
(2) We propose a novel intra-group credit rescaling method that refines self-generated credit using both successful and failed rollouts.
(3) We show that \methodname{} outperforms strong agentic RL baselines on long-horizon agentic tasks across multiple models at different scales.

\section{Preliminaries}
\label{sec:prelim}

\textbf{Long-horizon agentic tasks.} We formulate long-horizon tasks for LLM agents as follows. Given a query $Q$, an LLM agent $M_{\theta}$ interacts with an environment $\mathcal{E}$ and answers $Q$ by taking a sequence of actions over $T$ interaction turns, followed by a final-answer turn. At each interaction turn $t\in\{0,\dots,T-1\}$, the model generates a thinking trace $I_t$, takes an action $a_t\in\mathcal{A}$, and receives an observation $o_t\in\mathcal{O}$ from the environment. The agent's state is $s_t=(Q,\mathcal{H}_t)$, where $\mathcal{H}_t=\mathcal{H}_{0:t}$ and $\mathcal{H}_{k:t}=((I_k,a_k,o_k),\dots,(I_{t-1},a_{t-1},o_{t-1}))$ for $k\in\{0,\dots,t\}$, with $\mathcal{H}_{t:t}$ denoting the empty history. For example, in a tool-use setting, the model takes an action by issuing a call to an external tool with appropriate parameters. At the final turn $t=T$, the model generates a thinking trace $I_T$ and a final answer $a_T$, then terminates the interaction. The complete trajectory of $T+1$ turns is $\mathcal{T}=((I_0,a_0,o_0),\dots,(I_{T-1},a_{T-1},o_{T-1}),(I_T,a_T))$, which receives a binary reward $R(a_T,Q)$, abbreviated as $R\in\{0,1\}$.

\textbf{Multi-turn RL with GRPO.} We use GRPO \citep{shao2024deepseekmath} to train the model in this multi-turn RL setting. At each training step, the rollout policy $M_{\theta}$ generates $G$ trajectories $\{\mathcal{T}^j\}_{j=1}^G$ for the same query $Q$ through interactions with the environment, each receiving a reward $R^j$. Let $\Adv^{\,j}$ denote the advantage for trajectory $\mathcal{T}^j$; its detailed formulation can be found in Appendix~\ref{appx:formulation_details}.

Let $\mathcal{T}^{j}_{M_{\theta}}=((I^j_0,a^j_0),\dots,(I^j_{T^j},a^j_{T^j}))$ denote the sequence of model-generated thinking traces and actions, including the final answer, in trajectory $\mathcal{T}^j$, where $T^j$ is its number of interaction turns. We use $|\mathcal{T}^{j}_{M_{\theta}}|$ to denote the total number of model-generated tokens in this sequence. The advantage $\Adv^{\,j}$ is uniformly applied to every token in $\mathcal{T}^{j}_{M_{\theta}}$. Further details can be found in Appendix~\ref{appx:formulation_details}.

\textbf{Credit assignment.} GRPO with only outcome rewards provides coarse credit for a multi-turn decision chain. As the advantage $\Adv^{\,j}$ is broadcast uniformly to every model-generated token in the rollout, all turns share the same advantage.
Such reward sparsity becomes more problematic for long-horizon agentic tasks as the trajectory length grows, because a single outcome reward must supervise more turns \citep{feng2025groupingroup,zhang2026creditsurvey}. Providing better learning signals requires credit assignment, usually expressed through non-uniform advantage estimates. In this work, we focus on turn-level advantages $\Adv^{j}_{t}$.

\begin{figure*}[t]
     \centering

         \includegraphics[trim={0cm 4cm 0cm 3cm}, clip,width=\textwidth]{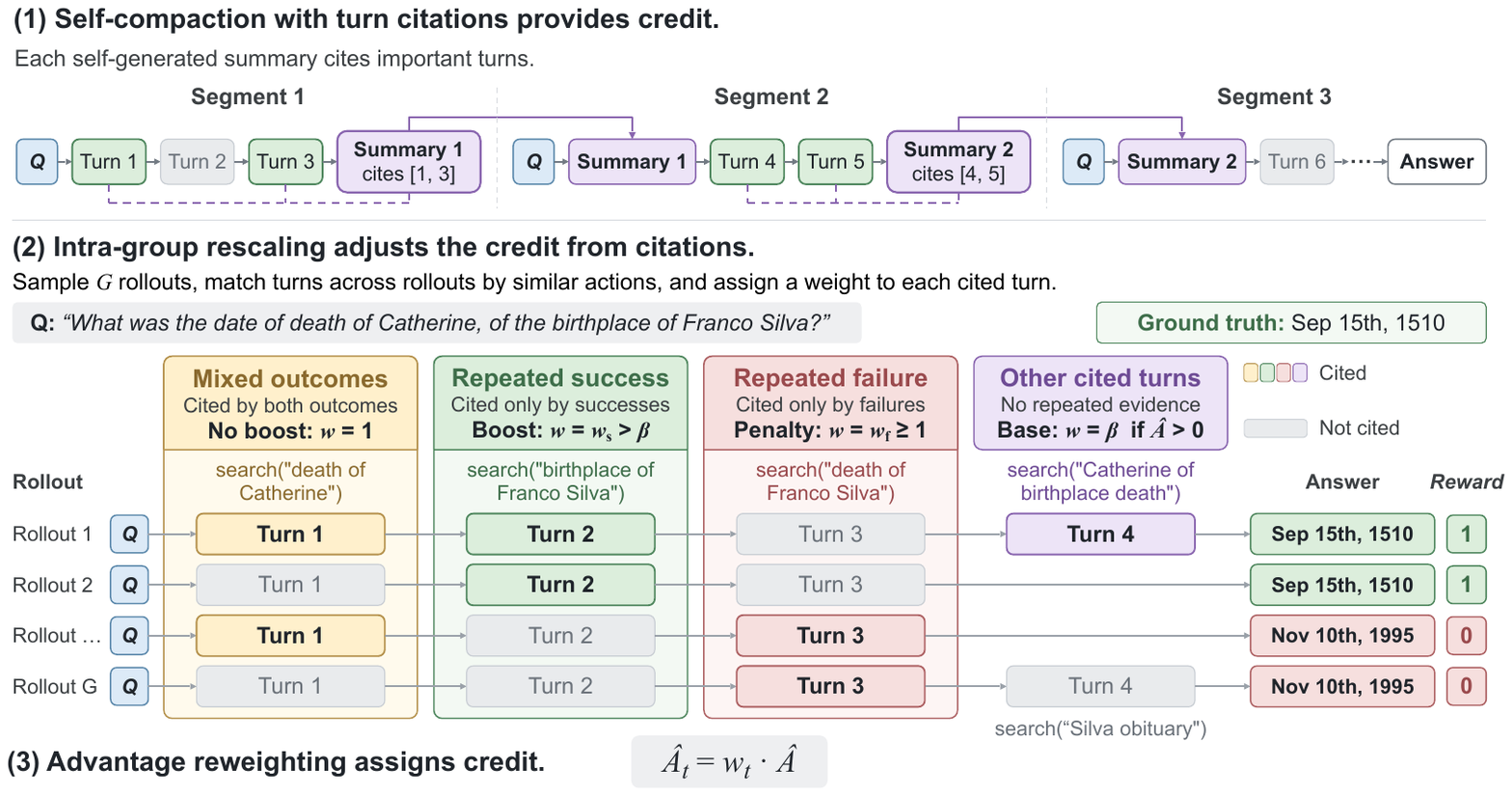}
         \caption{ Illustration of \methodname{}. (1) \methodname{} leverages self-compaction with turn citations for credit assignment. (2) Across rollouts for the same prompt, intra-group rescaling removes extra credit from cited turns that appear in both successful and failed rollouts, strengthens positive credit for turns unique to successes, and assigns stronger negative credit to turns that are only cited by failures. (3) The rescaled weights are used for advantage reweighting. }
         \label{fig:method_detail}

\end{figure*}

\section{Methods}
\label{sec:ours}

We present an approach that creates dense supervision signals from outcome rewards for long-horizon agentic RL using \textbf{self-generated credit with intra-group rescaling} via compaction. Our method, \methodname{}, combines three components: (1) the policy performs compaction by generating summaries of the interaction history, and the summaries cite important turns in the history as credit; (2) Intra-group rescaling refines this self-generated credit against the outcomes of other rollouts for the same prompt; (3) The rescaled credit is used to upweight the advantages of cited turns. Figure~\ref{fig:method_detail} illustrates \methodname{}.

\subsection{Self-Compaction with Turn Citations}
\label{sec:selfcite}

\textbf{Compaction-augmented RL.} Prior work introduces \emph{compaction} into RLVR for long-horizon agentic tasks to address issues caused by growing context \citep{li2026compactionrl,li2026selfcompact,lu2025supo}. Compaction replaces the interaction history in the context with a shorter summary, allowing generation to continue conditioned on the query and summary.

During training, compaction is triggered whenever the context reaches or exceeds $C$ tokens at turn $t$, where $C$ is a threshold chosen according to the effective context size of the LLM agent. When compaction is triggered, a summarizer $M_\phi$ produces a compact history $z_t$ from the current interaction context:

\begin{equation}
  \widetilde{\mathcal{H}}_{t':t} :=
  \begin{cases}
    \mathcal{H}_{t':t}, & |(Q,\mathcal{H}_{t':t})| < C, \\
    z_t,\quad z_t\sim M_\phi(\cdot\mid Q,\mathcal{H}_{t':t}),
      & |(Q,\mathcal{H}_{t':t})|\ge C.
  \end{cases}
  \label{eq:sum-transition}
\end{equation}

Here, $t'$ is the most recent turn at which compaction was triggered before turn $t$ ($t'=0$ if no compaction has been triggered yet), $\mathcal{H}_{t':t}$ is the context before compaction, and $|(Q,\mathcal{H}_{t':t})|$ denotes the context length. $\widetilde{\mathcal{H}}_{t':t}$ is the context after compaction: the previous interaction context is discarded and replaced by the summary; the model continues from the query and summary.

Compaction may be triggered multiple times as the context grows again after each summary. Let $N^j$ be the number of context-replacing summaries generated in trajectory $\mathcal{T}^j$. The trajectory is then divided into $N^j+1$ \emph{segments} $\{\mathcal{S}_i^j\}_{i=1}^{N^j+1}$, each conditioned on the query and the preceding summary if exists.

Following compaction-augmented policy optimization methods such as SUPO \citep{lu2025supo}, we extend the  GRPO objective to:

\begin{equation}
  \mathcal{J}_{\mathrm{SUM-RL}}(\theta)= \mathbb{E}\!\Biggl[
    \frac{1}{\sum_{j=1}^{G}|\mathcal{T}^{j}_{M_\theta}|}
    \sum_{j=1}^{G}\sum_{i=1}^{N^j+1}
    \sum_{\ell=1}^{|\mathcal{S}^j_{i,M_\theta}|}
    \min\Bigl(\rho^j_{i,\ell}\Adv^{\,j},\;
      \clip\bigl(\rho^j_{i,\ell},1-\varepsilon_{\mathrm{low}},
      1+\varepsilon_{\mathrm{high}}\bigr)\Adv^{\,j}
    \Bigr)
  \Biggr],
  \label{eq:supo}
\end{equation}

where $\mathcal{S}^j_{i,M_\theta}$ contains the trainable model-generated tokens of segment $i$.

Here, $\mathcal{T}^j_{M_\theta}$ includes all trainable policy-generated tokens in the rollout, including summary tokens, so $|\mathcal{T}^j_{M_\theta}|=\sum_{i=1}^{N^j+1}|\mathcal{S}^j_{i,M_\theta}|$. In other words, the objective computes one advantage per complete rollout trajectory and broadcasts it across the trainable tokens of all segments.

\textbf{Summary with citations.}
Following \citet{lu2025supo,li2026compactionrl,cursor2026composer2}, we use the policy model itself as the summarizer: $M_\phi=M_\theta$. During training and inference, when compaction is triggered, we prompt the policy model to summarize important information in the interaction history and cite important turns using their indices, producing a citation set $K\subseteq\{t',\dots,t-1\}$. During training, the model is penalized for generating an invalid summary or citation to enforce the required format. We also prompt the model to provide citations at the end of the rollout to obtain signals for the final segment. Details of the prompt and required format can be found in Appendix~\ref{appx:prompts}.

\subsection{Intra-group Rescaling}
\label{sec:credit-validation}

However, relying solely on self-generated credit or rewards can reinforce undesired behaviors; related work shows that optimizing model-generated feedback can lead to hacking and instability \citep{shafayat2025largereasoningmodelsselftrain,zhang2025freelunchrethinkinginternal,wang2025cream, wang2026detecting}. For example, instead of selectively citing important turns, the model can cite every turn in the context. Using such citations for credit assignment would then fail to provide selective learning signals.

To elicit better learning signals from self-generated citations, \methodname{} uses information already available in group-based RL without additional supervision or inference. For a given turn in a rollout, we examine similar turns across other rollouts in the same prompt group. A turn that appears more frequently across successful rollouts is more likely to be associated with success, and the same reasoning applies to failure. This provides within-group evidence for validating and refining the model's self-generated credit.

Given a query $Q$ and its group of $G$ rollout trajectories $\{\mathcal{T}^j\}_{j=1}^G$, let $\mathcal{C}$ and $\mathcal{F}$ be the sets of successful and failed rollouts, respectively. Let $B_t^j$ denote the $t$-th turn of trajectory $\mathcal{T}^j$. Following \citet{feng2025groupingroup}, we assume that some turns appear more than once and can be matched across rollouts. For example, in search-agent tasks, two search actions can be matched if the sets of documents they retrieve have substantial overlap. In addition, we assume that different turns in a rollout contribute independently to the final outcome. Details of the matching procedure and a discussion of this assumption are provided in Appendix~\ref{appx:turn_match}.

Let $n_+(B_t^j)$ and $n_-(B_t^j)$ denote the numbers of successful and failed rollouts, respectively, in the group that cite a turn matching $B_t^j$. Both numbers take the current rollout $\mathcal{T}^{j}$ into consideration. We use these counts to distinguish the following three critical scenarios for a cited turn: (1) \textbf{Repeated success}: a matching turn is cited by at least $\tau_+$ successful rollouts and no failed rollouts; (2) \textbf{Repeated failure}: a matching turn is cited by at least $\tau_-$ failed rollouts and no successful rollouts; and (3) \textbf{Mixed outcomes}: a matching turn is cited by at least one successful rollout and at least one failed rollout.

Then, for a cited turn $B_t^j$ with $\Adv^{\,j}\neq0$, we assign a scaling weight as follows:

\newcommand{\casenote}[1]{\quad\text{\footnotesize\color{gray}(#1)}}

\begin{equation}
  w_t^j :=
  \begin{cases}
    w_s   & \Adv^{\,j}>0,\ n_{-}(B_t^j)=0,\ \text{and } n_{+}(B_t^j)\ge\tau_{+}
            \casenote{Successes}\\[1pt]
    1     & \Adv^{\,j}>0\ \text{and } n_{-}(B_t^j)>0
            \casenote{Mixed}\\[2pt]
    \beta & \Adv^{\,j}>0,\ \text{otherwise}
            \casenote{Base credit}\\[2pt]
    w_f   & \Adv^{\,j}<0,\ n_{+}(B_t^j)=0,\ \text{and } n_{-}(B_t^j)\ge\tau_{-}
            \casenote{Failures}\\[2pt]
    1     & \Adv^{\,j}<0,\ \text{otherwise}. \\
  \end{cases}
  \label{eq:credit-validation}
\end{equation}

In other words, we refine credit assignment in these three scenarios through intra-group comparisons: (1) Repeated success: we apply a larger scaling weight $w_s>\beta$ to strengthen positive credit; (2) Repeated failure: we apply a scaling weight $w_f \geq 1$ to strengthen the negative advantage; and (3) Mixed outcomes: we remove the $\beta$ upweighting for successful rollouts because the effect of the cited turn is uncertain. For all other cited turns, we retain the base weight of $\beta$ for positive advantages and $1$ for negative advantages. For uncited turns $t \notin K$, we do not change the advantages, so $w^{j}_{t}=1$. The three scenarios are illustrated in Figure~\ref{fig:method_detail}. 

Intra-group rescaling is applied only to turns that can be matched across rollouts. Summary turns and final-answer turns are excluded from comparisons to preserve credit for summarization and answer commitment. Nonetheless, we show empirically in Section~\ref{sec:analysis} that rescaling can still be triggered frequently, as most agentic actions within a group are similar when conditioned on the same query. 

\subsection{Advantage Reweighting}
\label{sec:signaware}

We leverage the rescaled weights for \textbf{advantage reweighting}. We use the citation set to replace the uniform advantage broadcast in Equation~\ref{eq:supo} with turn-level weights. For each model-generated token $\ell$ in turn $t$ of trajectory $\mathcal{T}^j$, the token-level advantage is defined as follows:

\begin{equation}
  \Adv^{j}_{t,\ell}
  := \Adv^{\,j}\,w^{\,j}_{t}
  \label{eq:weighted-adv}
\end{equation}

In other words, all tokens in the same turn receive the same reweighted advantage using the assigned rescaled weight. In addition, in a successful rollout trajectory, summary turns and the final-answer turn are automatically treated as cited so that they receive credit without requiring explicit citations from the model.

\subsection{Annealing the Weights}
\label{sec:anneal}

The importance of exploration and exploitation varies across different stages of LLM post-training \citep{cui2025entropymechanismreinforcementlearning,cheng2025reasoningexplorationentropyperspective}. Since the rescaled weights can influence this balance, we design a learning curriculum with an annealing schedule for the repeated-failure penalty weight $w_f$. Stronger penalties early in training are intended to discourage repeatedly unsuccessful retrieval actions and stabilize search behaviors, while weaker penalties later allow more exploration.

\begin{table*}[t] \centering \small  \begin{NiceTabular}{@{}p{0.15\textwidth}p{0.15\textwidth}p{0.27\textwidth}p{0.30\textwidth}@{}} \toprule \Block{2-1}{\textbf{Method}} & \Block{2-1}{\textbf{Long Context Management}} & \Block[c]{1-2}{\textbf{Credit Assignment}} & \\ \cmidrule(lr){3-4} & & \textbf{Form of credits} & \textbf{Source of credits} \\ \midrule GRPO & --& --& -- \\ GiGPO & --& Advantage reweighting & Successful and failed rollouts \\ SUPO & Summary & --& -- \\ ECHO & Turn Selection & Turn masking & Successful rollouts only \\ \midrule \textbf{\methodname{}} & Summary & Advantage reweighting & Successful and failed rollouts \\ \bottomrule \end{NiceTabular} 

\caption{Key differences between \methodname{} and baseline methods. 
\methodname{} leverages self-generated summaries during training to provide credit assignment via advantage reweighting that is grounded in both successful and failed rollouts via intra-group rescaling.
} 
\label{tab:methods_diff}
\end{table*}

\section{Experimental Setup}
\label{sec:exp}

\subsection{Setup}
\label{sec:setup}

\paragraph{Task} We evaluate our method on the following long-horizon multi-hop QA and deep research tasks: (1) \textbf{BrowseComp-Plus} \citep{chen2025browsecompplusfairtransparentevaluation}: LLM agents interact with a search engine and retrieve documents to solve multi-hop questions; (2) Deep research tasks: we evaluate on \textbf{GAIA} \citep{mialon2024gaia} and \textbf{xBench-DeepSearch} \citep{chen2025xbenchtrackingagentsproductivity}. These tasks include real-world questions requiring multi-step reasoning, web browsing, and tool use. For GAIA, we use its text-only subset. 


\paragraph{Models} We train Qwen3-8B, Qwen3-32B \citep{yang2025qwen3technicalreport}, and Gemma4-E4B \citep{gemmateam2026gemma4technicalreport} with thinking mode. 

\paragraph{Training} For BrowseComp-Plus, we follow \citet{xie2026echo} and split the dataset into 747 training examples for RL training and 83 test examples for evaluation. We use the BrowseComp-Plus static corpus and index as the search backend. For deep research tasks, we train on 1K deep research QA pairs from OpenResearcher \citep{li2026openresearcherfullyopenpipeline} and use the OpenResearcher static browser as the tool use harness and search backend during training. For efficient training, we use a fully asynchronous rollout with VeRL \citep{sheng2024hybridflow} and Megatron \citep{megatron-lm} frameworks. For \methodname{}, we use $\tau_+=\tau_-=2$ in all experiments. Details can be found under Appendix~\ref{appx:training_configs}.



\subsection{Baselines}
\label{sec:baselines}
We compare against the following zero-shot and agentic RL baselines. The main differences between \methodname{} and these baselines can be found in Table~\ref{tab:methods_diff}. (1) \textbf{Zero-shot} evaluation of the base model does not use context management; (2) \textbf{GRPO} is applied to the multi-turn setting without any context management or turn-level credit assignment; (3) \textbf{GiGPO} \citep{feng2025groupingroup} modifies group-based RL with turn-level credit assignment by re-grouping the same rollouts based on action states that are matched by repetitive actions. Turns sharing an identical environment observation are matched into groups, and within each group the action receives a standardized additive advantage adjustment; (4) \textbf{SUPO} \citep{lu2025supo} is a compaction-augmented RL baseline that implements a rolling summarization with a free-form summary and optimizes the uniform-credit objective. There is no citation instruction in the summarization prompt and no turn-level credit assignment; (5) \textbf{ECHO} \citep{xie2026echo} replaces free-form summarization in summarization-augmented RL with explicit turn selection: at each summarization, the policy emits a selection over its turn history, and the selected turn memories are carried into the next context in place of a summary. It routes token-level credit to the turns selected by the policy in successful rollouts. 

\subsection{Evaluation Protocol}
\label{sec:protocol}

For BrowseComp-Plus, we follow \citet{chen2025browsecompplusfairtransparentevaluation,xie2026echo} and use the Qwen3-Embedding-8B \citep{zhang2025qwen3embeddingadvancingtext} retriever and the BrowseComp-Plus static corpus index as the search engine during evaluation. We report LLM-as-a-judge accuracy, prompting a separate Qwen3-32B judge to determine whether the model-generated answer is semantically equivalent to the ground-truth answer.
For deep research tasks, we follow \citet{li2026openresearcherfullyopenpipeline} to use open web search API instead of the static engine during evaluation. We report LLM-as-a-judge accuracy as well with the rubric from \citet{gupta2026deepsearchqabridgingcomprehensivenessgap}. For both tasks, we use temperature = 0.7 for decoding.

\section{Results}
\label{sec:main}

\begin{table}[t]
    \centering
    \begin{minipage}[t]{0.55\textwidth}\vspace{0pt}
    \centering
    \setlength{\tabcolsep}{3pt}
    \begin{NiceTabular}{lcccc}
    \CodeBefore
    \rectanglecolor{metabg}{8-2}{8-5}
    \Body
    \toprule
    & \multicolumn{2}{c}{\textit{Qwen3-32B}}
    & \multicolumn{2}{c}{\textit{Qwen3-8B}} \\
    \cmidrule(lr){2-3}\cmidrule(lr){4-5}
    & Acc. & \# Turns & Acc. & \# Turns \\
    \midrule
    Zero-shot
        & 10.8 & -
        & 3.6 & - \\
    GRPO
        & 33.3\textsubscript{$\pm$1.4} & 13.2\textsubscript{$\pm$0.4}
        & 27.7\textsubscript{$\pm$2.4} & 10.8\textsubscript{$\pm$0.2} \\
    GiGPO
        & 38.2\textsubscript{$\pm$0.8} & 10.9\textsubscript{$\pm$0.2}
        & 26.1\textsubscript{$\pm$1.7} & 6.7\textsubscript{$\pm$0.1} \\
    SUPO
        & 36.9\textsubscript{$\pm$4.6} & 40.8\textsubscript{$\pm$1.3}
        & 32.1\textsubscript{$\pm$2.5} & 13.4\textsubscript{$\pm$0.6} \\
    ECHO
        & 41.4\textsubscript{$\pm$2.5} & 45.8\textsubscript{$\pm$2.7}
        & 32.1\textsubscript{$\pm$3.0} & 62.6\textsubscript{$\pm$2.1} \\
    \midrule
    \textbf{\methodname{}}
        & \textbf{47.0}\textsubscript{$\pm$2.1} & 39.4\textsubscript{$\pm$2.4}
        & \textbf{40.6}\textsubscript{$\pm$0.7} & 37.9\textsubscript{$\pm$1.9} \\
    \bottomrule
    \end{NiceTabular}
    \caption{Test accuracy (Pass@1) and average number of turns on BrowseComp-Plus. Results are reported over 3 seeds. \methodname{} outperforms strong RL baselines across model scales.}
    \label{tab:main}
    \end{minipage}\hfill
    \begin{minipage}[t]{0.41\textwidth}\vspace{0pt}
    \centering
    \setlength{\tabcolsep}{3pt}
    \begin{NiceTabular}{lcc}
    \CodeBefore
    \rectanglecolor{metabg}{7-2}{7-3}
    \Body
    \toprule
    & GAIA & xBench-DeepSearch \\
    \midrule
    & Acc. & Acc. \\
    \midrule
    Zero-shot
        & 23.4 & 4.0 \\
    GRPO
        & 34.3\textsubscript{$\pm$0.6} & 11.3\textsubscript{$\pm$2.9} \\
    GiGPO
        & 13.3\textsubscript{$\pm$0.6} & 12.0\textsubscript{$\pm$1.0} \\
    SUPO
        & 27.2\textsubscript{$\pm$2.4} & 6.3\textsubscript{$\pm$1.5} \\
    ECHO
        & 32.0\textsubscript{$\pm$2.6} & 10.0\textsubscript{$\pm$1.7} \\
    \midrule
    \textbf{\methodname{}}
        & \textbf{39.2}\textsubscript{$\pm$1.5} & \textbf{16.7}\textsubscript{$\pm$1.5} \\
    \bottomrule
    \end{NiceTabular}
    \caption{Test accuracy (Pass@1) on GAIA and xBench-DeepSearch with Gemma4-E4B. Results are reported over 3 seeds. \methodname{} outperforms strong RL baselines on both deep research benchmarks.}
    \label{tab:main_dr}
    \end{minipage}
\end{table}

\begin{figure*}[t]
     \centering

         \includegraphics[width=\textwidth]{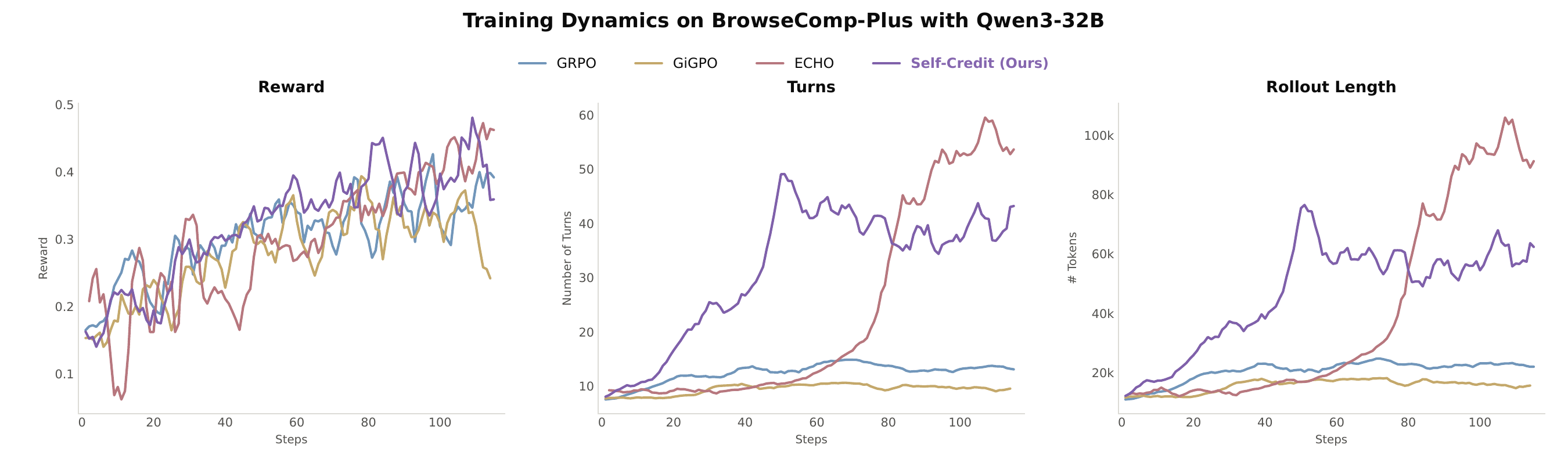}
         \caption{Training dynamics on BrowseComp-Plus on Qwen3-32B models of RL baselines with different credit assignment methods. \methodname{} learns stably across steps without excessive use of tool calls.}
         \label{fig:bcp_32b_training_dynamics}

\end{figure*}

\label{sec:bcp-results}

\textbf{\methodname{} outperforms strong RL baselines across settings.} Table~\ref{tab:main}
reports test accuracy on BrowseComp-Plus. \methodname{} outperforms all baselines at both Qwen3 model scales and beats the strongest baseline by $5.6$ points at 32B and $8.5$ points at 8B. Table~\ref{tab:main_dr} summarizes results on deep research tasks by training a Gemma4-E4B model. On GAIA, \methodname{} achieves the highest accuracy and outperforms the strongest baseline by $4.9$ points. On xBench-DeepSearch, \methodname{} outperforms the best baseline by $4.7$ points. We also show that \methodname{} trains models that generalize better to OOD shorter horizon search QA datasets in Appendix~\ref{appx:ood_gen}. These results demonstrate consistent gains across the evaluated model families and benchmarks with \methodname{}.

\textbf{\methodname{} learns efficient long-horizon tool use.} Training dynamics of different methods on BrowseComp-Plus shows the strength of \methodname{} from balancing depth and efficiency of tool use for long horizon tasks. Figure~\ref{fig:bcp_32b_training_dynamics} shows later in training, \methodname{} typically uses fewer turns than ECHO does, while attaining high training reward. GRPO and GiGPO generate much shorter trajectories and exhibit lower training reward. This suggests that simply minimizing or maximizing interaction length for long-horizon tasks is insufficient~\citep{yin2025livemcp101}. Together, these results show that \methodname{} supports sufficient exploration to gather useful evidence while limiting excessive interaction.

\section{Analysis}
\label{sec:analysis}

\subsection{Ablations}
\label{sec:ablations}

\begin{table}[t]
    \centering
    \setlength{\tabcolsep}{4pt}
   \begin{NiceTabular}{@{}lcc@{}}
    \CodeBefore
    \rectanglecolor{metabg}{3-1}{3-3}
    \rectanglecolor{metabg}{7-1}{7-3}
    \Body
    \toprule
                                                & \textit{Qwen3-32B} & \textit{Qwen3-8B} \\
    \midrule
     \textbf{\methodname{}}        & \textbf{47.0} & \textbf{40.6} \\                         
     Ablations on \emph{Form of Credits} && \\
                                     --w/o Intra-group rescaling     & 41.0  & 28.9  \\
                                     --w/o Penalties for repeated failures & 34.9 & 26.5 \\
                                     --w/o Rescaling for mixed outcomes   & 41.0 & 31.3  \\
                                       Ablations on  \emph{Source of Credits} && \\
                                     --w/o Self-generated citations & 42.2 & 27.7 \\
                                     --w/ Random citations &  38.6  & 32.5 \\
    \bottomrule
    \end{NiceTabular}
    \caption{Ablations of \methodname{} on BrowseComp-Plus. Intra-group rescaling, penalties for repeated failures, and self-generated credits are important for the downstream performance using \methodname{}. }
    \label{tab:ablations}
\end{table}

To isolate the importance of different components of \methodname{}, we conduct an ablation analysis with the following variants: (1) \textbf{without intra-group rescaling}: we train models to summarize with citations and \emph{upweight} the advantages of all cited turns in rollouts with positive advantages; (2) \textbf{without penalties for repeated failures}; (3) \textbf{without rescaling for mixed outcomes}; (4) \textbf{without self-generated citations}: we apply intra-group rescaling and advantage reweighting to \emph{all turns} instead of only the model-cited turns; (5) \textbf{with random citations}: we apply intra-group rescaling and advantage reweighting to a random set of turns in every segment, averaged over 3 seeds. 

\begin{table}[t]
    \centering
    \small
    \begin{NiceTabular}{@{}ll*{4}{>{\centering\arraybackslash}p{0.13\textwidth}}@{}}
    \CodeBefore
    \rectanglecolor{metabg}{6-2}{6-6}
    \rectanglecolor{metabg}{11-2}{11-6}
    \Body
    \toprule
                                            &                               & Gold Searched (\%) & Gold Opened (\%)   & Answer Recall (\%)   & Evidence Recall (\%) \\
    \midrule
    \multirow{5}{*}{\textit{Qwen3-32B}}     & GRPO                          & 42.2           & 25.3          & 33.9          & 29.5          \\
                                            & GiGPO                         & 42.2           & 12.0          & 25.1          & 25.3          \\
                                            & SUPO                          & 53.0           & 25.3          & 43.2          & 39.6          \\
                                            & ECHO                          & 62.7           & 30.1          & 51.6          & \textbf{48.3} \\
                                            & \textbf{\methodname{}} & \textbf{66.3}  & \textbf{49.4} & \textbf{53.8} & 47.0          \\
    \midrule
    \multirow{5}{*}{\textit{Qwen3-8B}}      & GRPO                          & 42.2           & 25.3          & 28.0          & 25.1          \\
                                            & GiGPO                         & 34.9           & 0.0           & 19.2          & 16.9          \\
                                            & SUPO                          & 42.2           & 9.6           & 30.4          & 27.8          \\
                                            & ECHO                          & 48.2           & 27.7          & 36.0          & 31.7          \\
                                            & \textbf{\methodname{}} & \textbf{50.6}  & \textbf{32.5} & \textbf{36.7} & \textbf{32.0} \\
    \bottomrule
    \end{NiceTabular}
    \caption{Analysis of the quality of tool use on BrowseComp-Plus for different training methods. Gold Searched and Gold Opened measure the percentage of questions for which at least one gold document appears among the search results or is opened, respectively. Answer Recall and Evidence Recall measure the fractions of documents containing the gold answer or supporting evidence that are found through search or page opening.}
    \label{tab:tool_use_quality}
\end{table}

Table~\ref{tab:ablations} summarizes the performance of these variants on BrowseComp-Plus. Among the intra-group components, rescaling repeated turns in failed rollouts is especially important for both model scales. Removing it reduces accuracy from $47.0\%$ to $34.9\%$ for Qwen3-32B and from $40.6\%$ to $26.5\%$ for Qwen3-8B. We observe that without penalties for repeated turns cited by failed rollouts, Qwen3-8B model learns to search excessively, likely caused by over-optimizing for the credit boost, and fails to learn to commit to an answer. Moreover, applying credit assignment to all turns or randomly selected turns reduces Qwen3-8B accuracy to $27.7\%$ and $32.5\%$, respectively, showing that self-generated citations provide a better basis for credit assignment than using arbitrary turns. 

Overall, these results show that the different components of \methodname{} work together: both self-generated citations and intra-group rescaling, especially over repeated failures, are important for achieving the best performance.

\subsection{Quality of Credit Assignment}
\label{sec:citation_analysis}

\begin{table}[t]
    \centering
    \begin{NiceTabular}{llccc}
    \CodeBefore
    \rectanglecolor{metabg}{5-2}{5-5}
    \rectanglecolor{metabg}{9-2}{9-5}
    \Body
    \toprule
                                            &                        & Opened        & Searched      & Judge F1        \\
    \midrule
    \multirow{4}{*}{\textit{Qwen3-32B}}     & Random                 & 18.2          & 32.8          & 44.7          \\
                                            & Cite all               & 21.2          & 43.1          & 60.1          \\
                                            & ECHO                   & \textbf{40.7} & 39.7          & 60.6          \\
                                            & \textbf{\methodname{}} & 36.0          & \textbf{59.7} & \textbf{71.7} \\
    \midrule
    \multirow{4}{*}{\textit{Qwen3-8B}}      & Random                 & 22.6          & 32.6          & 41.3          \\
                                            & Cite all               & 26.9          & 41.4          & 53.9          \\
                                            & ECHO                   & 27.7          & 37.4          & 62.4          \\
                                            & \textbf{\methodname{}} & \textbf{45.7} & \textbf{58.1} & \textbf{69.3} \\
    \bottomrule
    \end{NiceTabular}
    \caption{Quality of citations generated by different credit assignment methods on BrowseComp-Plus. \emph{Opened} and \emph{Searched} report F1 (\%) between cited turns and turns whose \emph{Open} or \emph{Search} actions find gold documents. \emph{Judge F1} reports F1 (\%) against turns judged relevant by a stronger LLM judge.}
    \label{tab:citation_f1}
\end{table}

\begin{figure}[t]
     \centering

         \includegraphics[width=\linewidth]{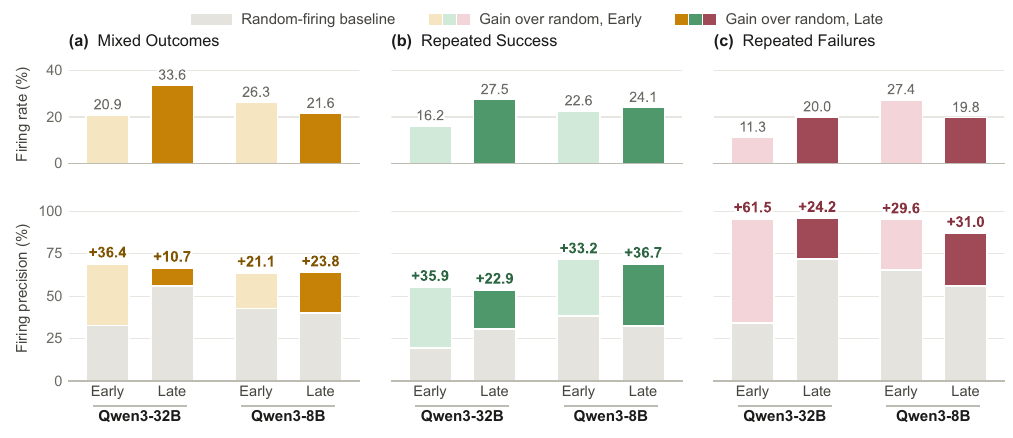}
        \caption{Firing rate and precision of \methodname{} intra-group rescaling across the three rules at different training stages on BrowseComp-Plus. \emph{Firing rate} denotes how frequently intra-group rescaling is applied to a cited turn, and \emph{firing precision} indicates whether rescaling gives the correct credit to the cited turns. Intra-group rescaling is triggered selectively during training for different models.}
         \label{fig:validation_firing}

\end{figure}

\textbf{\methodname{} improves agentic tool use.}  We evaluate the quality of agentic tool use of models trained using different methods on BrowseComp-Plus, with a focus on two main types of tool use in this task: \emph{Search} and \emph{Open}. Details of the evaluation metrics are in Appendix~\ref{appx:analysis_citation}. Table~\ref{tab:tool_use_quality} summarizes the results. Models trained with \methodname{} achieve higher Gold Searched, Gold Opened, and Answer Recall scores than the baselines at both model sizes. 

\textbf{\methodname{} generates better citations.} To assess the quality of citations generated by different methods, we evaluate whether the cited turns perform actions that find gold documents for the agentic tasks. On BrowseComp-Plus, we use \emph{Open} or \emph{Search} actions that find gold documents as a proxy for critical actions, and we compute the F1 score of cited turns against the turns containing such actions. We also compute F1 against the important turn judged by Claude-4.6-Sonnet. Details of the evaluation metrics are in Appendix~\ref{appx:analysis_citation}. Table~\ref{tab:citation_f1} reports the results for ECHO and \methodname{}; for reference, we introduce a random citation baseline that cites a subset of all turns in the rollout uniformly at random, and a baseline that cites all turns. \methodname{} achieves higher citation quality than ECHO and baselines for multiple metrics, especially better at identifying critical search turns. This indicates \methodname{} trains a better policy at finding the important turns in its generated rollouts, leading to a more reliable source for credit assignment.

\textbf{Intra-group rescaling is applied selectively and accurately.} We evaluate how frequently the intra-group rescaling is applied to cited turns during the early (0--60 steps) and late (60--115 steps) training stages on BrowseComp-Plus. We also report the precision of rescaling by checking whether it gives extra credit to actions that are related to the gold document, and reduces extra credit or give penalties for other less relevant actions. Details of the evaluation metrics are in Appendix~\ref{appx:analysis_citation}. Figure~\ref{fig:validation_firing} shows that intra-group rescaling is triggered selectively. Compared with a random selection baseline, \methodname{} provides higher quality learning signals with high precision of crediting the good turns and downweighting or penalizing the other turns.

%


\section{Related Work}
\label{sec:related}

\paragraph{Agentic RL for long-horizon tasks.}
Reinforcement learning (RL) is widely used to post-train LLM agents \citep{jin2025searchr,singh2025agenticreasoningtoolintegration,qian2025toolrl,wang2025ragenunderstandingselfevolutionllm},
building on policy-gradient methods such as PPO
\citep{schulman2017ppo} and GRPO \citep{shao2024deepseekmath}. In long-horizon tasks~\citep{sun2026agentslastexam}, growing interaction histories can exceed the context window, while sparse outcome rewards provide limited
information about the contribution of individual actions \citep{wang_multi_turn,tang2026visuallygroundedselfreflectionvisionlanguage}. Recent work addresses these challenges by using inference-time methods \citep{xu2025amem,ye2026dyscodynamicattentionscalingdecoding}, incorporating context management systems \citep{li2026selfcompact,zhang2026memgen,kang2026acon, zhou2026mem} or self-compaction capability \citep{lu2025supo,li2026compactionrl,cursor2026composer2} into the policy model through RL.
Our work leverages self-compaction for context management during training for long horizon tasks.

\paragraph{Credit assignment under sparse outcome rewards.}
Prior work derives finer grained credit given sparse outcome rewards from several sources, including learned critics \citep{zhou2025sweetrltrainingmultiturnllm,shen2026carlcriticalityawareagenticreinforcement,zhang2025criticsearchfinegrainedcreditassignment,shi2026odysseusscalingvlms100}, process reward models \citep{zhang2026trca,zeng2025reinforcing,liu2026agentic}, hindsight information conditioned on answers \citep{tan2026hcapo,zhang2026trca} and sampling \citep{choudhury2025agentprm,zong2026at2po}, but these methods usually require additional rollouts or hindsight information not available in intermediate steps. Other methods reuse sampled rollouts to compare returns at similar states  \citep{feng2025groupingroup,he2026hierarchyofgroups}, or leverage confidence of the policy model \citep{wang2025igpo,dong2025agenticentropybalancedpolicyoptimization,wang2026tpo}. Compared with prior work, our method combines signals that exist in standard group-based RL methods with self-compaction and rescales them using group-level statistics. 


\section{Conclusion}
\label{section:conclusion}

Training long-horizon LLM agents with RLVR is challenging because outcome rewards provide very sparse learning signals over long trajectories. We introduced \methodname{}, an end-to-end agentic RL method that creates dense supervision signals from a single sparse outcome reward through \textbf{self-generated credit with intra-group rescaling}. The policy model generates summaries with citations to critical turns during compaction, and these citations serve as self-generated credit. \methodname{} then refines this credit with intra-group rescaling, which compares cited actions across successful and failed rollouts for the same prompt, and uses the rescaled credit weights to selectively strengthen the advantages of tokens in model-cited turns. \methodname{} outperforms the strongest agentic RL baselines on challenging search QA and deep research tasks across models. These results suggest that self-generated signals can be utilized for effective credit assignment in long-horizon agentic RL.

\section*{Acknowledgments}
This work was partially supported by the Sloan Foundation, NSF CAREER Award IIS-2145280, NSF grant IIS-2433071, and the NSF under Cooperative Agreement 2421782 and the Simons Foundation grant MPS-AI-00010515 awarded to the NSF-Simons AI Institute for Cosmic Origins -- CosmicAI, \url{https://www.cosmicai.org/}.



\clearpage
\newpage
\bibliographystyle{assets/plainnat}
\bibliography{paper}

\clearpage
\newpage
\beginappendix
\section{Formulation Details}
\label{appx:formulation_details}
\subsection{GRPO Formulation}
The advantage in GRPO is formulated as follows:

\begin{equation}
  \Adv^{\,j}
  = \frac{R^j-\mean\big(\{R^{j'}\}_{j'=1}^{G}\big)}
         {\std\big(\{R^{j'}\}_{j'=1}^{G}\big)+\epsilon}, \quad j\in\{1,\dots,G\},
  \label{eq:grpo-adv}
\end{equation}
where $\epsilon>0$ is a small constant for numerical stability.

The GRPO objective is defined as 
\begin{equation}
  \mathcal{J}_{\mathrm{GRPO}}(\theta)= \mathbb{E}\!\Biggl[
    \frac{1}{\sum_{j=1}^{G}|\mathcal{T}^{j}_{M_{\theta}}|}
    \sum_{j=1}^{G}\sum_{\ell=1}^{|\mathcal{T}^{j}_{M_{\theta}}|}
    \min\Bigl(\rho^j_{\ell}\Adv^{\,j},\quad
      \clip\bigl(\rho^j_{\ell},1-\varepsilon_{\mathrm{low}},
      1+\varepsilon_{\mathrm{high}}\bigr)\Adv^{\,j}
    \Bigr)
  \Biggr],
  \label{eq:grpo}
\end{equation}
where $\rho^j_{\ell}$ is the token-level importance ratio for the $\ell$-th token in $\mathcal{T}^{j}_{M_{\theta}}$. 

\paragraph{Importance sampling ratio in GRPO.} Importance sampling ratio is the ratio of its probability under $M_\theta$ to that under $M_{\theta_{\mathrm{old}}}$, both conditioned on the context used to generate that token. Although token-level importance ratios and clipping can produce different gradient contributions, they do not provide turn-specific credit based on each action's contribution to the outcome.

\section{Implementation Details}

\label{appx:impl_details}
\subsection{Prompts and Instructions}
\label{appx:prompts}
The summarization instruction used in
our experiments is the following.

\begin{prompt}[title={\thetcbcounter{} Summarization Instruction}, label=example:sum_inst]
As the LAST line INSIDE the \texttt{<summary>} block, cite ONLY the few assistant turns whose findings were decisive for solving the task---usually 2 to 5 turns, NEVER every turn---written
exactly in this format: \texttt{the critical turns are: <comma-separated turn numbers>}. \\

Pick the real turn numbers yourself based on the interaction; do NOT copy the placeholder and do NOT list every turn. Number the assistant turns starting from 1 for the first assistant tool-calling turn shown in the context above; do not count tool responses, any earlier summary block, or this instruction.

\end{prompt}

\paragraph{Citation format.} We introduce simple format gating on the generated citations: the citation line cannot be empty, and we strip out-of-range citations. An invalid citation line leads to a format error and makes the entire rollout receive a reward of 0.

\paragraph{Summary for the final segment.} Because compaction is only triggered at certain thresholds, turns in a short last segment might not get a chance to be cited, creating a systematic bias in the credit signal. Therefore, when a rollout trajectory is complete with a final answer, we force the model to generate an auxiliary summary of the final segment. This auxiliary summary is not a context-replacing summary and is not included in $N^j$.

\subsection{Turn Matching}
\label{appx:turn_match}
\paragraph{Matching turns across rollouts by actions.}
Intra-group rescaling requires deciding when a cited turn in one rollout performs the same action as a turn in other rollouts for the same prompt. As defined in Section~\ref{sec:prelim}, a turn consists of thinking trace, an agentic action, and an observation from the environment. For the setting of long horizon agentic tasks, we focus on \emph{actions} to match turns because LLM agents learn to perform the right actions during training. In the two settings below, instead of comparing the text of tool calls, which rarely repeats verbatim across rollouts, we identify each action by the \emph{evidence it retrieves} from the environment. 

\paragraph{BrowseComp-Plus.}
On BrowseComp-Plus, the agent interacts with a fixed document corpus through three tools: (1) \texttt{Search} takes a natural-language query and returns a ranked list of retrieved documents, each of which has a document identifier (docid); (2) \texttt{open\_page} takes a docid and returns the full text of that document; (3) \texttt{finish} submits the final answer and ends the interaction. We do not match turns with \texttt{finish} calls. For turns that leverage the other 2 types of tools, we use the following matching rules: (1) A \texttt{search} call is represented by the set of docids it returned. Within each prompt group, we match search turns by the overlap between the sets of retrieved docids. Two turns are matched if the Jaccard similarity of their sets of retrieved docids is at least $0.6$; (2) For \texttt{open\_page}, we match turns by the docid they opened , and two \texttt{open\_page} turns match only if they opened the same document. 

\paragraph{Deep research tasks.}
For deep research tasks, the agent browses a fixed web corpus with OpenResearch harness during training through four tools: (1) \texttt{search} takes a natural-language query and returns the top 20 results as links; (2) \texttt{open} takes a link id or cursor from an earlier observation and returns the page content with line numbers; (3) \texttt{find} takes a text pattern and returns the lines in an opened page that match it; (4) \texttt{finish} submits the final answer and ends the interaction. Unlike on BrowseComp-Plus, the agent refers to pages through link ids (e.g., URLs) and cursors (location on the website) that are local to each rollout, making matching harder. Therefore, we identify each document by its docid in the OpenResearcher corpus, which the retriever returns as metadata hidden from the agent, instead of by links or cursors. As in BrowseComp-Plus, we do not match turns with \texttt{finish} calls, and we match \texttt{search} turns by the Jaccard similarity of their retrieved docids (at least $0.6$) and \texttt{open} turns by the docid they opened. For \texttt{find}, two turns match if they found adjacent lines in the same document.

\paragraph{Turn matching for other tasks.}
The matching rules above are instances of a general recipe: a task-specific function matches turns by the observations from those turns. Intra-group rescaling only needs these matches, so it can be applied unchanged to other agentic tasks once such a function is defined. This recipe is related to anchor-state grouping in GiGPO \citep{feng2025groupingroup}, which groups steps across trajectories that start from the same environment state, using exact matching or a similarity threshold over the state text. The difference lies in what is matched: GiGPO compares the states in which actions are taken, while \methodname{} compares the observations that cited actions obtain, which remains informative in open-ended environments where states rarely repeat. For example, in embodied tasks such as ALFWorld \citep{shridhar2021alfworld}, a turn can be described by the object and receptacle it acts on (e.g., an object picked by the agent), together with whether the action succeeded. 

\paragraph{Independence Assumption} In this work, we assume that different turns in a rollout contribute independently to the outcome, which is an approximation: an open action depends on the search that surfaced the document, and a late turn may only succeed because of evidence gathered earlier. We adopt the approximation because it keeps the comparison to counts over rollouts rather than over trajectory prefixes, which would rarely match.

\section{Training Configurations}
\label{appx:training_configs}
\subsection{General Hyperparameters}

Following \citet{xie2026echo,li2026openresearcherfullyopenpipeline}, we use the same training configuration on BrowseComp-Plus and deep research tasks for all training methods, including \methodname{}:
\begin{itemize}
    \item \textbf{Optimization.} We use a learning rate of $1\times10^{-6}$ with a constant schedule and asymmetric clipping with $\epsilon_{\text{low}}=0.20$ and $\epsilon_{\text{high}}=0.28$. The loss is averaged over tokens, with no entropy bonus, no KL penalty in either the loss or the reward, and no critic.
    \item \textbf{Batching.} Each training step uses 256 trajectories, split into 8 optimizer updates. We use $16$ prompts per step and $16$ rollouts per prompt.
    
    \item \textbf{Sequence lengths.} The maximum prompt length is 4096 tokens and the maximum response length is 32768 tokens. The threshold for compaction is 16384 tokens.
    
    \item \textbf{Rollouts.} Rollouts are generated asynchronously in a multi-turn agent loop, with tool calls in the Hermes format. Tool responses are truncated to 16K characters. The maximum number of assistant turns is 100.

For BrowseComp-Plus, we allow the model to use 5 parallel tool calls and train for 115 steps, following the settings used by \citet{xie2026echo}. For deep research tasks, we allow the model to use 1 parallel tool call. We train for 100 steps. All training runs are conducted on 8x8 NVIDIA B200 GPU.
    
\end{itemize}

\subsection{Hyperparameters of Baselines}
We use the official implementations of the baseline methods.

\paragraph{ECHO.} We set turn-selection top-k to $5$ and keep $3$ recent turns verbatim.

\paragraph{GiGPO.} We use discount $\gamma = 0.95$. Turns are grouped by similarity of the raw observation texts, as implemented by the original GiGPO paper.

\subsection{Hyperparameters of \methodname{}}

For all models on both datasets, we set maximum number of summary rounds to 10, $\beta=1.5$, $w_{s} = 2.5$, and $w_{f} = 1.5$, annealed to $w_{f} = 1.0$ linearly.

\section{OOD Generalization}
\label{appx:ood_gen}

\begin{table*}[t]
    \centering
    \small
    \begin{NiceTabular}{lcccc|c}
    \toprule
    & HotpotQA & MuSiQue & FRAMES & Bamboogle & \textbf{Avg.} \\
    \midrule
                 GRPO      & 64.5 & 29.4 & 48.4 & 39.2 & 45.4 \\
                     GiGPO      & 63.2 & 31.6 & \textbf{51.7} & 32.0 & 44.6 \\
SUPO      & 63.7 & 30.6 & 47.8 & 35.2 & 44.3 \\
                                     ECHO      & 63.4 & \textbf{31.8} & 49.2 & 33.6 & 44.5 \\
    \midrule
    \textbf{\methodname{}}                   & \textbf{65.1} & 28.0 & \textbf{51.7} & \textbf{41.6} & \textbf{46.6} \\
    \bottomrule
    \end{NiceTabular}
    \caption{Zero-shot generalization on out-of-distribution search-augmented QA tasks.}
    \label{tab:ood}
\end{table*}
\subsection{Out-of-distribution Generalization}

\label{sec:ood-results}

To evaluate the out-of-distribution transfer of different methods to shorter horizon tasks
after training on long horizon, we evaluate zero-shot transfer of the Qwen3-32B policies to four out-of-distribution
search-augmented QA benchmarks, following prior work \citep{jin2025searchr,xie2026echo}. As shown in Table~\ref{tab:ood}, \methodname{} achieves the
highest average accuracy, $46.6\%$, compared with $45.4\%$ for GRPO, $44.5\%$ for ECHO, and $44.3\%$ for SUPO. This demonstrates that tool use behaviors learned with long horizon training with \methodname{} generalize to shorter tasks as well. 

\section{Details of Analysis}

\label{appx:analysis_citation}

\paragraph{Tool-use quality metrics.} BrowseComp-Plus annotates each question with gold documents, which contain the answer, and a larger set of evidence documents, which are needed to support it. \emph{Gold Searched} and \emph{Gold Opened} are the percentages of questions for which at least one gold document appeared in any search result or was opened, respectively. \emph{Answer Recall} is the fraction of a question's gold documents that the policy either retrieved or opened, averaged over questions, and \emph{Evidence Recall} is the same quantity computed over evidence documents.

\paragraph{Citation quality metrics in Table~\ref{tab:citation_f1}.}(1) \emph{Opened} and \emph{Searched}. We measure whether the turns a model cites contain the evidence needed for the answer, using the gold documents that BrowseComp-Plus annotates in the dataset. An Open or Search turn is considered important if it opened a gold document (\emph{Opened}) or if one of its search results was a gold document (\emph{Searched}). We compute the F1 of the cited turns against these turns, excluding the final answer turn. (2) \emph{LLM Judge.} A useful exploratory turn may not find any gold documents but still inform the model about what to do next. While these turns still provide learning signals, the metrics above might overlook such turns. Therefore, we use \emph{LLM Judge} F1 to score model citations against turns that Claude Sonnet 4.6 labels as decisive or as finding new relevant information, which does not rely on gold annotations. We use the LLM-as-judge prompt from \citet{xu2026triageroletypedcreditassignment}.

\paragraph{Rule firing and precision in Figure~\ref{fig:validation_firing}.} We measure how often each intra-group rescaling rule scenario is applied during training and whether it is applied to the right turns. The \emph{firing rate} is the fraction of cited turns on which a rule is applied. The \emph{firing precision} checks fired turns against turns that find the gold documents, i.e., turns that either opened a gold document or if one of its search results was a gold document. For the repeated success scenario, the firing precision is the share of fired turns that find gold documents. For repeated failure and mixed outcome, its precision is the share of fired turns that do not find gold documents. 

\section{Licenses} \label{appx:licenses}
We use the following publicly available datasets from prior works with open licenses.
\paragraph{BrowseComp-Plus} We access the BrowseComp-Plus data that uses the MIT license and data is available at: 
\url{https://huggingface.co/datasets/Tevatron/browsecomp-plus}.

\paragraph{GAIA} We access GAIA data that uses the CC BY 4.0 license and data is available at: 
\url{https://huggingface.co/datasets/gaia-benchmark/GAIA}.

\paragraph{xBench-DeepSearch} We access the data that uses the MIT license and data is available at: 
\url{https://huggingface.co/datasets/xbench/DeepSearch}.

\section{Output Example} 
\label{appx:example_output}

We show an example output from the Qwen3-8B model trained with \methodname{} on BrowseComp-Plus. Thinking traces and tool outputs are truncated to their first two sentences; summaries and citation lines are shown in full. At the end of each context window the model writes a summary and cites the assistant turns it found decisive; the next window restarts from the system prompt, the query and that summary.

\newpage

\begin{figure*}

\begin{imagebox}[]{Example: \methodname{} rollout on \emph{BrowseComp-Plus} Test Question}%

\textbf{Query:} There’s an American rapper and songwriter who was born in the 1990s and in October. He joined a gang when he was 14. He has a “Libra” zodiac sign. Between the years 2015-2020 (inclusive), he released a viral hit song, which was from one of his album's that was released between the years 2015-2019 (inclusive) and that song featured another American rapper whose father spent 15 years in the penitentiary, as of January 21, 2016. Can you tell me the name of that viral song?\\[2pt]
\textbf{Ground-truth answer:} Mando

\end{imagebox}

\begin{modelsection}[]{Segment 1, Turn 1}

[User] System prompt (research protocol and tool definitions; omitted) followed by the query above.\\

[Assistant]

{\small\ttfamily\raggedright\noindent
\mbox{}\textless{}think\textgreater{}\\
\mbox{}Okay, let me try to figure out how to approach this question. The user is asking for the name of\\
\mbox{}~~a viral song by an American rapper and songwriter born in the 1990s, who joined a gang at 14\\
\mbox{}~~and has a Libra zodiac sign. ...\\
\mbox{}\textless{}/think\textgreater{}\\
\mbox{}\textless{}tool\_call\textgreater{}\\
\mbox{}\{{\char34}name{\char34}: {\char34}search{\char34}, {\char34}arguments{\char34}: \{{\char34}query{\char34}: {\char34}American rapper born 1990s joined gang at 14 Libra\\
\mbox{}~~zodiac sign viral song 2015-2020 featuring rapper with father 15 years prison 2016{\char34}, {\char34}topk{\char34}:\\
\mbox{}~~10\}\}\\
\mbox{}\textless{}/tool\_call\textgreater{}\par}

\par\noindent [Tool] \emph{(search, 8 documents returned; Doc 1 shown)}

{\small\ttfamily\raggedright\noindent
\mbox{}\textless{}tool\_response\textgreater{}\\
\mbox{}Doc 1 (Title: Hip-Hop Artists By Zodiac Signs) {[}docid: 88327{]}\\
\mbox{}The entirety of hip-hop is pretty much obsessed with Zodiac signs. We can go as far as back as\\
\mbox{}~~OutKast\textquotesingle{}s 1998 album Aquemini, the title for which combined Big Boi\textquotesingle{}s astrological sign with\\
\mbox{}~~that of Andre 3000\textquotesingle{}s. ...\\
\mbox{}\textless{}/tool\_response\textgreater{}\par}

\end{modelsection}

\begin{modelsection}[]{Segment 1, Turn 2}

[Assistant]

{\small\ttfamily\raggedright\noindent
\mbox{}\textless{}think\textgreater{}\\
\mbox{}Okay, let me try to figure out the answer here. The user is asking for the name of a viral song\\
\mbox{}~~by an American rapper born in the 1990s, who joined a gang at 14, has a Libra zodiac sign, and\\
\mbox{}~~released a song between 2015-2020 featuring another rapper whose father spent 15 years in\\
\mbox{}~~prison as of January 2016. ...\\
\mbox{}\textless{}/think\textgreater{}\\
\mbox{}\textless{}tool\_call\textgreater{}\\
\mbox{}\{{\char34}name{\char34}: {\char34}open\_page{\char34}, {\char34}arguments{\char34}: \{{\char34}docid{\char34}: {\char34}50738{\char34}\}\}\\
\mbox{}\textless{}/tool\_call\textgreater{}\par}

\par\noindent [Tool] \emph{(open\_page)}

{\small\ttfamily\raggedright\noindent
\mbox{}\textless{}tool\_response\textgreater{}\\
\mbox{}Document {[}docid: 50738{]}: EXCLUSIVE: \$tupid Young on Joining Cambodian Crips at 14, Getting\\
\mbox{}~~{\char34}Jumped In{\char34}\\
\mbox{}In this VladTV exclusive, \$tupid Young talks about the age he got put on to gang culture. The\\
\mbox{}~~rapper broke down the politics and cycles of gang banging, and even mentioned the last\\
\mbox{}~~situation he went through in 2015 where he lost a close friend after getting ...\\
\mbox{}\textless{}/tool\_response\textgreater{}\par}

\end{modelsection}

\caption{Example output of \methodname{} model on a BrowseComp-Plus question.} \label{fig:selfcite_bcp_exp}

\end{figure*}

\clearpage

\begin{figure*}

\begin{modelsection}[]{Segment 1, Turn 3}

[Assistant]

{\small\ttfamily\raggedright\noindent
\mbox{}\textless{}think\textgreater{}\\
\mbox{}Okay, let me see. The user is asking for the name of a viral song by an American rapper with\\
\mbox{}~~specific background details. ...\\
\mbox{}\textless{}/think\textgreater{}\\
\mbox{}\textless{}tool\_call\textgreater{}\\
\mbox{}\{{\char34}name{\char34}: {\char34}open\_page{\char34}, {\char34}arguments{\char34}: \{{\char34}docid{\char34}: {\char34}43835{\char34}\}\}\\
\mbox{}\textless{}/tool\_call\textgreater{}\par}

\par\noindent [Tool] \emph{(open\_page)}

{\small\ttfamily\raggedright\noindent
\mbox{}\textless{}tool\_response\textgreater{}\\
\mbox{}Document {[}docid: 43835{]}: \$tupid Young Age, Birthday, Zodiac Sign and Birth Chart\\
\mbox{}\$tupid Young, born Alex Pham on October 3, 1992, is an Asian-American rapper, singer,\\
\mbox{}~~songwriter, and record executive. He rose to fame in 2018 with the hit single \textquotesingle{}Mando,\textquotesingle{}\\
\mbox{}~~featuring Mozzy, and achieved further success with the song \textquotesingle{}Trust Nobody\textquotesingle{} the same ...\\
\mbox{}\textless{}/tool\_response\textgreater{}\par}

\end{modelsection}

\begin{modelsection}[]{Segment 1, Turn 4}

[Assistant]

{\small\ttfamily\raggedright\noindent
\mbox{}\textless{}think\textgreater{}\\
\mbox{}Okay, let me try to figure out the answer here. The user is asking for the name of a viral song\\
\mbox{}~~by an American rapper with specific background details. ...\\
\mbox{}\textless{}/think\textgreater{}\\
\mbox{}\textless{}tool\_call\textgreater{}\\
\mbox{}\{{\char34}name{\char34}: {\char34}open\_page{\char34}, {\char34}arguments{\char34}: \{{\char34}docid{\char34}: {\char34}52452{\char34}\}\}\\
\mbox{}\textless{}/tool\_call\textgreater{}\par}

\par\noindent [Tool] \emph{(open\_page)}

{\small\ttfamily\raggedright\noindent
\mbox{}\textless{}tool\_response\textgreater{}\\
\mbox{}Document {[}docid: 52452{]}: Lil Durk\\
\mbox{}Durk Derrick Banks (October 19, 1992), better known by his stage name Lil Durk, is an American\\
\mbox{}~~rapper, singer, and songwriter. Early life{[}{]} Durk Derrick Banks was born on October 19, 1992\\
\mbox{}~~in Englewood, Illinois. ...\\
\mbox{}\textless{}/tool\_response\textgreater{}\par}

\end{modelsection}

\begin{modelsection}[]{Segment 1, Turn 5}

[Assistant]

{\small\ttfamily\raggedright\noindent
\mbox{}\textless{}think\textgreater{}\\
\mbox{}Okay, let me try to figure out the answer here. The user is asking for the name of a viral song\\
\mbox{}~~by an American rapper with specific background details. ...\\
\mbox{}\textless{}/think\textgreater{}\\
\mbox{}\textless{}tool\_call\textgreater{}\\
\mbox{}\{{\char34}name{\char34}: {\char34}open\_page{\char34}, {\char34}arguments{\char34}: \{{\char34}docid{\char34}: {\char34}30656{\char34}\}\}\\
\mbox{}\textless{}/tool\_call\textgreater{}\par}

\par\noindent [Tool] \emph{(open\_page)}

{\small\ttfamily\raggedright\noindent
\mbox{}\textless{}tool\_response\textgreater{}\\
\mbox{}Document {[}docid: 30656{]}: Lil Durk - Wikipedia\\
\mbox{}name: Lil Durk birth\_name: Durk Derrick Banks alias: The Voice, Smurk, Mustafa Abdul Malak Lil\\
\mbox{}~~Durk New Muslim Name Instagram birth\_date: 19 October 1992 Lil Durk – Universal Music France\\
\mbox{}~~birth\_place: Chicago, Illinois, U.S.March 14, 2016 Lil Durk Takes ...\\
\mbox{}\textless{}/tool\_response\textgreater{}\par}

\end{modelsection}

\caption{Figure~\ref{fig:selfcite_bcp_exp} continuation.}

\end{figure*}

\clearpage

\begin{figure*}

\begin{modelsection}[]{Segment 1, Summary and turn citation}

\par\noindent [User]

{\small\ttfamily\raggedright\noindent
\mbox{}System: You are running low on context. Summarize everything important (the original question,\\
\mbox{}~~key verified facts with their source docids, useful search/tool results so far, dead ends, and\\
\mbox{}~~the exact next step) so you can continue solving from this summary alone. ...\par}

[Assistant]

{\small\ttfamily\raggedright\noindent
\mbox{}\textless{}summary\textgreater{}\\
\mbox{}From the search results, we have a few key points:\\
\mbox{}\\
\mbox{}1. **\$tupid Young** (born Alex Pham, October 3, 1992) is an Asian-American rapper with a Libra\\
\mbox{}~~zodiac sign. He joined a gang at a young age and has a history of incarceration. His 2018 hit\\
\mbox{}~~{\char34}Mando{\char34} features Mozzy, but there\textquotesingle{}s no mention of Mozzy\textquotesingle{}s father\textquotesingle{}s prison history.\\
\mbox{}\\
\mbox{}2. **Lil Durk** (Durk Derrick Banks, born October 19, 1992) has a similar background of joining\\
\mbox{}~~a gang at a young age and his father\textquotesingle{}s incarceration. However, the user\textquotesingle{}s query specifies the\\
\mbox{}~~father spent 15 years in prison as of January 2016, which doesn\textquotesingle{}t align with Lil Durk\textquotesingle{}s\\
\mbox{}~~father\textquotesingle{}s situation.\\
\mbox{}\\
\mbox{}3. The user\textquotesingle{}s query mentions a song from 2015-2020 featuring another rapper whose father spent\\
\mbox{}~~15 years in prison. The search results don\textquotesingle{}t provide explicit information on the featured\\
\mbox{}~~rapper\textquotesingle{}s father\textquotesingle{}s prison history, but \$tupid Young\textquotesingle{}s {\char34}Mando{\char34} (2018) fits the timeline and\\
\mbox{}~~zodiac sign. However, without confirmation on the featured rapper\textquotesingle{}s father\textquotesingle{}s prison sentence,\\
\mbox{}~~there\textquotesingle{}s uncertainty.\\
\mbox{}\\
\mbox{}The next step is to verify if Mozzy (featured on {\char34}Mando{\char34}) has a father with a 15-year prison\\
\mbox{}~~sentence. If not, the answer might be {\char34}Mando{\char34} by \$tupid Young, but further confirmation is\\
\mbox{}~~needed.\\
\mbox{}\textcolor{green!40!black}{\textbf{The critical turns are: 1, 2, 3.}}\par}
\mbox{}\textless{}/summary\textgreater{}\\


\par\noindent\emph{Citation check (not shown to the model): cited turns 1, 2, 3; turns that opened or retrieved a gold/evidence document: 1, 2, 3.}

\end{modelsection}

\begin{modelsection}[]{Segment 2, context}

[User] System prompt and query (repeated), then the summary of Segment 1 shown above (without its citation line), followed by: ``The above is a summary of your previous interaction history. Please continue solving the problem based on this context. You may call tools as needed.''

\end{modelsection}

\begin{modelsection}[]{Segment 2, Turn 1}

[Assistant]

{\small\ttfamily\raggedright\noindent
\mbox{}\textless{}think\textgreater{}\\
\mbox{}Okay, let me try to figure out the answer here. The user is asking for the name of a viral song\\
\mbox{}~~by an American rapper born in the 1990s, October, with a Libra zodiac sign. ...\\
\mbox{}\textless{}/think\textgreater{}\\
\mbox{}\textless{}tool\_call\textgreater{}\\
\mbox{}\{{\char34}name{\char34}: {\char34}search{\char34}, {\char34}arguments{\char34}: \{{\char34}query{\char34}: {\char34}Mozzy father prison 15 years{\char34}, {\char34}topk{\char34}: 5\}\}\\
\mbox{}\textless{}/tool\_call\textgreater{}\par}

\par\noindent [Tool] \emph{(search, 5 documents returned; Doc 1 shown)}

{\small\ttfamily\raggedright\noindent
\mbox{}\textless{}tool\_response\textgreater{}\\
\mbox{}Doc 1 (Title: Mozzy - Wikipedia) {[}docid: 74122{]}\\
\mbox{}name: Mozzy birth\_name: Timothy Patterson ACE RepertorySaponara Michael Mozzy Arrested for Gun\\
\mbox{}~~Possession in Las Vegas: Report 22 May 2019 Billboard 9 July 2018 alias: Lil Tim birth\_date:\\
\mbox{}~~24 06 1987 birth\_place: Oak Park, Sacramento, California, U.S. ...\\
\mbox{}\textless{}/tool\_response\textgreater{}\par}

\end{modelsection}

\caption{Figure~\ref{fig:selfcite_bcp_exp} continuation.}

\end{figure*}

\clearpage

\begin{figure*}

\begin{modelsection}[]{Segment 2, Turn 2}

[Assistant]

{\small\ttfamily\raggedright\noindent
\mbox{}\textless{}think\textgreater{}\\
\mbox{}Okay, let me see. The user is asking for the name of a viral song by an American rapper born in\\
\mbox{}~~the 1990s, October, with a Libra zodiac sign. ...\\
\mbox{}\textless{}/think\textgreater{}\\
\mbox{}\textless{}tool\_call\textgreater{}\\
\mbox{}\{{\char34}name{\char34}: {\char34}open\_page{\char34}, {\char34}arguments{\char34}: \{{\char34}docid{\char34}: {\char34}74122{\char34}\}\}\\
\mbox{}\textless{}/tool\_call\textgreater{}\par}

\par\noindent [Tool] \emph{(open\_page)}

{\small\ttfamily\raggedright\noindent
\mbox{}\textless{}tool\_response\textgreater{}\\
\mbox{}Document {[}docid: 74122{]}: Mozzy - Wikipedia\\
\mbox{}name: Mozzy birth\_name: Timothy Patterson ACE RepertorySaponara Michael Mozzy Arrested for Gun\\
\mbox{}~~Possession in Las Vegas: Report 22 May 2019 Billboard 9 July 2018 alias: Lil Tim birth\_date:\\
\mbox{}~~24 06 1987 birth\_place: Oak Park, Sacramento, California, U.S. ...\\
\mbox{}\textless{}/tool\_response\textgreater{}\par}

\end{modelsection}

\begin{modelsection}[]{Segment 2, Turn 3}

[Assistant]

{\small\ttfamily\raggedright\noindent
\mbox{}\textless{}think\textgreater{}\\
\mbox{}Okay, let me check the documents again. The user is asking for the name of a viral song by an\\
\mbox{}~~American rapper born in the 1990s, October, with a Libra zodiac sign. ...\\
\mbox{}\textless{}/think\textgreater{}\\
\mbox{}\textless{}tool\_call\textgreater{}\\
\mbox{}\{{\char34}name{\char34}: {\char34}search{\char34}, {\char34}arguments{\char34}: \{{\char34}query{\char34}: {\char34}Mando song Mozzy{\char34}, {\char34}topk{\char34}: 5\}\}\\
\mbox{}\textless{}/tool\_call\textgreater{}\par}

\par\noindent [Tool] \emph{(search, 5 documents returned; Doc 1 shown)}

{\small\ttfamily\raggedright\noindent
\mbox{}\textless{}tool\_response\textgreater{}\\
\mbox{}Doc 1 (Title: Mando (feat. Mozzy)) {[}docid: 9640{]}\\
\mbox{}1 Track November 17, 2017 Afficials / EMPIRE Afficials / EMPIRE Release Mando (feat. Mozzy)\\
\mbox{}~~Featuring \$tupid Young Track Name Artist BPM Key Time Genre Label Release Date Tracks 1 97 BPM\\
\mbox{}~~B Minor 2:51 2017-11-17 Add to playlist Create a playlist Add to cart ...\\
\mbox{}\textless{}/tool\_response\textgreater{}\par}

\end{modelsection}

\begin{modelsection}[]{Segment 2, Turn 4}

[Assistant]

{\small\ttfamily\raggedright\noindent
\mbox{}\textless{}think\textgreater{}\\
\mbox{}Okay, let me check the documents again. The user is asking for the name of a viral song by an\\
\mbox{}~~American rapper born in the 1990s, October, with a Libra zodiac sign. ...\\
\mbox{}\textless{}/think\textgreater{}\\
\mbox{}\textless{}tool\_call\textgreater{}\\
\mbox{}\{{\char34}name{\char34}: {\char34}finish{\char34}, {\char34}arguments{\char34}: \{{\char34}answer{\char34}: {\char34}The viral song is \textbackslash{}{\char34}Mando\textbackslash{}{\char34} by \$tupid Young\\
\mbox{}~~featuring Mozzy.{\char34}, {\char34}explanation{\char34}: {\char34}Mozzy\textquotesingle{}s father, Marcus Mosley Sr., served 15 years in\\
\mbox{}~~prison for drug trafficking in the 1990s {[}docid:74122{]}. Mozzy is featured on \$tupid Young\textquotesingle{}s\\
\mbox{}~~2018 song \textbackslash{}{\char34}Mando,\textbackslash{}{\char34} which fits the 2015-2020 release window {[}docid:9640{]}. ...{\char34}, {\char34}confidence{\char34}:\\
\mbox{}~~{\char34}95\%{\char34}\}\}\\
\mbox{}\textless{}/tool\_call\textgreater{}\par}

\par\noindent [Tool]

{\small\ttfamily\raggedright\noindent
\mbox{}\textless{}tool\_response\textgreater{}\\
\mbox{}Answer submitted: The viral song is {\char34}Mando{\char34} by \$tupid Young featuring Mozzy. \textbar{} Explanation:\\
\mbox{}~~Mozzy\textquotesingle{}s father, Marcus Mosley Sr., served 15 years in prison for drug trafficking in the 1990s\\
\mbox{}~~{[}docid:74122{]}. Mozzy is featured on \$tupid Young\textquotesingle{}s 2018 song {\char34}Mando,{\char34} ...\\
\mbox{}\textless{}/tool\_response\textgreater{}\par}

\end{modelsection}

\caption{Figure~\ref{fig:selfcite_bcp_exp} continuation.}

\end{figure*}

\clearpage

\begin{figure*}

\begin{modelsection}[]{Final Summary}

\par\noindent [User]

{\small\ttfamily\raggedright\noindent
\mbox{}System: Summarize everything important (the original question,\\
\mbox{}~~key verified facts with their source docids, useful search/tool results, dead ends). ...\par}

[Assistant]

{\small\ttfamily\raggedright\noindent
\mbox{}\textless{}summary\textgreater{}\\
\mbox{}From the provided documents, we can confirm that Mozzy (Timothy Patterson) is the rapper in\\
\mbox{}~~question. His father, Marcus Mosley Sr., spent 15 years in prison for drug trafficking in the\\
\mbox{}~~1990s, aligning with the user\textquotesingle{}s description. Mozzy is featured on \$tupid Young\textquotesingle{}s 2018 song\\
\mbox{}~~{\char34}Mando,{\char34} which fits the timeline of a viral hit between 2015-2020. \$tupid Young, born in\\
\mbox{}~~October 1992, has a Libra zodiac sign and joined a gang at 14, matching the user\textquotesingle{}s details.\\
\mbox{}~~The song {\char34}Mando{\char34} is the viral hit in question, featuring Mozzy. The previous search results\\
\mbox{}~~and the current document confirm this information. Therefore, the answer is {\char34}Mando{\char34} by \$tupid\\
\mbox{}~~Young featuring Mozzy.\\
\mbox{}\textcolor{green!40!black}{\textbf{The critical turns are: 1, 2.}}\par}
\mbox{}\textless{}/summary\textgreater{}\\


\par\noindent\emph{Citation check (not shown to the model): cited turns 1, 2; turns that opened or retrieved a gold/evidence document: 1, 2, 3.}

\end{modelsection}

\begin{modelsection}[]{Final answer}

\textbf{Answer:} The viral song is "Mando" by \$tupid Young featuring Mozzy.\\

\textbf{Judge:} Correct (ground truth: Mando)

\end{modelsection}

\caption{Figure~\ref{fig:selfcite_bcp_exp} continuation.}

\end{figure*}

\clearpage

\end{document}